\documentclass[letterpaper]{article} 
\usepackage{aaai2026}  
\usepackage{times}  
\usepackage{helvet}  
\usepackage{courier}  
\usepackage[hyphens]{url}  
\usepackage{graphicx} 
\usepackage{natbib}  
\usepackage{caption} 
\DeclareUnicodeCharacter{030C}{\v{}}

\usepackage{amsmath}
\usepackage{courier}
\usepackage[table]{xcolor}
\usepackage{titletoc}
\usepackage{tocloft}
\usepackage{lipsum}
\usepackage{url}
\usepackage{subcaption}
\usepackage{listings}
\usepackage[nolist]{acronym}
\usepackage{amsmath}
\usepackage[font={small}]{caption}
\usepackage[export]{adjustbox}
\usepackage{amssymb}
\usepackage{bm}
\usepackage{booktabs}
\usepackage{color}
\usepackage{dsfont}
\usepackage{esvect}
\usepackage{float}
\usepackage{import}
\usepackage{mathtools}
\usepackage{multirow}
\usepackage{multicol}
\usepackage{outline}
\usepackage{paralist}
\usepackage{siunitx}
\usepackage{stackengine}
\usepackage{stmaryrd}
\usepackage{systeme}
\usepackage{soul}
\usepackage{url}
\usepackage{tikz}
\usetikzlibrary{tikzmark}
\usetikzlibrary{calc}

\usepackage{enumitem}
\usepackage[ruled,vlined]{algorithm2e}
\usepackage{scalerel}
\usepackage{makecell}

\definecolor{codeblue}{rgb}{0.25,0.5,0.5}
\definecolor{myblue}{rgb}{0.92, 0.97, 0.85}

\definecolor{mygreen}{rgb}{0.92, 1.0, 0.92}
\definecolor{myred}{rgb}{1, 0.9, 0.9}
\definecolor{mygray}{gray}{0.95}
\definecolor{mydarkblue}{rgb}{0,0.08,1}
\definecolor{mydarkred}{rgb}{0.8,0.02,0.02}
\definecolor{mydarkorange}{rgb}{0.40,0.2,0.02}
\definecolor{mypurple}{RGB}{111,0,255}
\definecolor{mygold}{rgb}{0.75,0.6,0.12}
\definecolor{mydarkgray}{rgb}{0.66, 0.66, 0.66}
\definecolor{mydarkgreen}{rgb}{0.02,0.6,0.02}
\definecolor{mygray}{gray}{0.9}
\definecolor{keynotegreen}{rgb}{0.04,0.52,0}
\definecolor{keynoteyellow}{rgb}{1,0.68,0}
\definecolor{LightCyan}{rgb}{0.88,1,1}
\definecolor{tabfirst}{rgb}{1, 0.7, 0.7}
\definecolor{tabsecond}{rgb}{1, 0.85, 0.7} 
\definecolor{tabthird}{rgb}{1, 1, 0.7} 
\definecolor{rbtred}{rgb}{255, 0, 0}

\newlength\paramargin
\newlength\figmargin
\newlength\subfigmargin
\newlength\secmargin
\newlength\subsecmargin
\newlength\tabmargin
\newlength\eqmargin

\SetKwComment{Comment}{\color{green!50!black}\# }{}

\SetKwProg{Function}{def}{:}{}

\SetKwProg{For}{for}{:}{}
\SetKwProg{If}{if}{:}{}

\def\method{CorrectNav\xspace}

\usepackage{newfloat}
\usepackage{listings}
\DeclareCaptionStyle{ruled}{labelfont=normalfont,labelsep=colon,strut=off} 
\floatstyle{ruled}
\newfloat{listing}{tb}{lst}{}
\floatname{listing}{Listing}
\title{CorrectNav: Self-Correction Flywheel Empowers \\ Vision-Language-Action Navigation Model}

\author {
    Zhuoyuan Yu\textsuperscript{*\rm 1\rm 2},
    Yuxing Long\textsuperscript{*\dag\rm 1\rm 2},
    Zihan Yang\textsuperscript{\rm 1\rm 2},
    Chengyan Zeng\textsuperscript{\rm 2}, \\
    Hongwei Fan\textsuperscript{{\rm 1\rm 2}},
    Jiyao Zhang\textsuperscript{{\rm 1\rm 2}},
    Hao Dong\textsuperscript{{\ddag\rm 1\rm 2}}
}
\affiliations {
    \textsuperscript{\rm 1}CFCS, School of Computer Science, Peking University,
    \textsuperscript{\rm 2}PKU-Agibot Lab\\
     *Equal contribution, \dag~Project Leader, \ddag~Corresponding author\\
     \url{https://correctnav.github.io}
}
\usepackage{bibentry}

\makeatletter
\@ifundefined{isChecklistMainFile}{
  \newif\ifreproStandalone
  \reproStandalonetrue
}{
  \newif\ifreproStandalone
  \reproStandalonefalse
}
\makeatother

\ifreproStandalone
\begin{document}

\fi
\setlength{\leftmargini}{20pt}
\makeatletter\def\@listi{\leftmargin\leftmargini \topsep .5em \parsep .5em \itemsep .5em}
\def\@listii{\leftmargin\leftmarginii \labelwidth\leftmarginii \advance\labelwidth-\labelsep \topsep .4em \parsep .4em \itemsep .4em}
\def\@listiii{\leftmargin\leftmarginiii \labelwidth\leftmarginiii \advance\labelwidth-\labelsep \topsep .4em \parsep .4em \itemsep .4em}\makeatother

\setcounter{secnumdepth}{0}
\renewcommand\thesubsection{\arabic{subsection}}
\renewcommand\labelenumi{\thesubsection.\arabic{enumi}}

\newcounter{checksubsection}
\newcounter{checkitem}[checksubsection]

\newcommand{\checksubsection}[1]{%
  \refstepcounter{checksubsection}%
  \paragraph{\arabic{checksubsection}. #1}%
  \setcounter{checkitem}{0}%
}

\newcommand{\checkitem}{%
  \refstepcounter{checkitem}%
  \item[\arabic{checksubsection}.\arabic{checkitem}.]%
}
\newcommand{\question}[2]{\normalcolor\checkitem #1 #2 \color{blue}}
\newcommand{\ifyespoints}[1]{\makebox[0pt][l]{\hspace{-15pt}\normalcolor #1}}

\twocolumn[{%
\renewcommand\twocolumn[1][]{#1}%

\maketitle

\begin{center}
    \centering 
    \vspace{-2em}
    \includegraphics[width=0.87\linewidth]{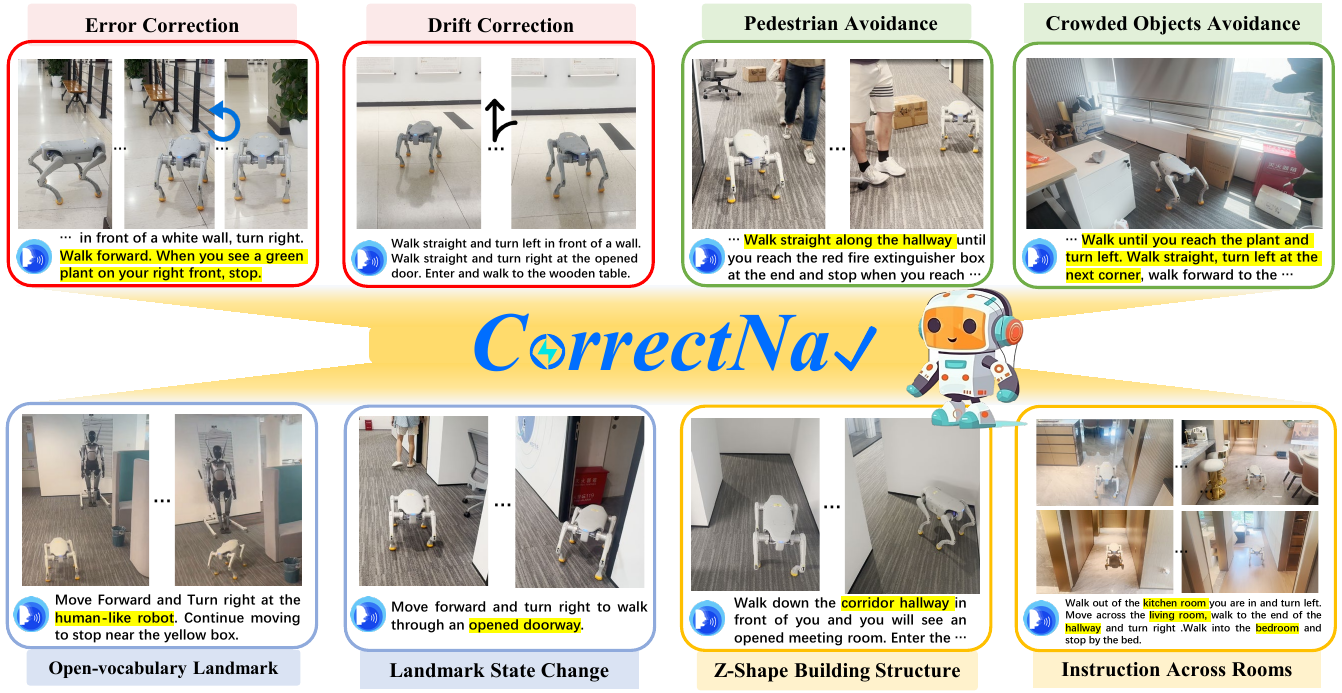}
    \captionof{figure}{\textbf{Diverse Capabilities of \method.} The model takes only monocular RGB video
    and language instructions as inputs, predicting navigation actions. Empowered by the Self-correction Flywheel post-training, \method not only maintains outstanding \textbf{multimodal reasoning} (Blue), but also displays improved \textbf{deviation correction} (Red), \textbf{obstacle avoidance} (Green), and \textbf{complex action execution} (Yellow).}
    \label{fig:teasor}
\end{center}
}]
 
\begin{abstract}
Existing vision-and-language navigation models often deviate from the correct trajectory when executing instructions. However, these models lack effective error correction capability, hindering their recovery from errors. To address this challenge, we propose Self-correction Flywheel, a novel post-training paradigm. Instead of considering the model’s error trajectories on the training set as a drawback, our paradigm emphasizes their significance as a valuable data source. We have developed a method to identify deviations in these error trajectories and devised innovative techniques to automatically generate self-correction data for perception and action. These self-correction data serve as fuel to power the model’s continued training. The brilliance of our paradigm is revealed when we re-evaluate the model on the training set, uncovering new error trajectories. At this time, the self-correction flywheel begins to spin. Through multiple flywheel iterations, we progressively enhance our monocular RGB-based VLA navigation model \method. Experiments on R2R-CE and RxR-CE benchmarks show \method achieves new state-of-the-art success rates of 65.1\% and 69.3\%, surpassing prior best VLA navigation models by 8.2\% and 16.4\%. Real robot tests in various indoor and outdoor environments demonstrate \method's superior capability of error correction, dynamic obstacle avoidance, and long instruction following.
\end{abstract}

\section{Introduction}
In the Vision-and-Language Navigation (VLN) task, users control the robot to move to desired locations in unexplored environments via natural language instructions, like “Move forward and turn right into the living room to wait near the sofa.” Due to its user-friendly interaction characteristic, VLN becomes a fundamental capability essential to embodied intelligence and attracts widespread research interest. During the navigation process, models inevitably predict wrong movement actions, causing the robot to deviate from the correct path. These deviations often produce misalignment between the environment and the instructions. Taking the above instruction as an example. If the robot directly turns right at the current position rather than moving forward first, it will enter the kitchen and cannot locate the sofa. At this time, the robot easily gets confused about such misalignment and fails to reach the destination.

Existing VLN models mainly focus on enhancing visual perception and multimodal reasoning capabilities by improving feature representation~\cite{an2024etpnav,hong2023learning} or increasing training data~\cite{zhang2024navid,uninavid}. They aim to enable the model to navigate correctly as much as possible in every step. However, the reality turned out to be different from expectations. Only several imperfect step-wise predictions can accumulate significant deviation from the correct path and ultimately cause failure. The absence of self-correction ability makes previous VLN models struggle to recover from mistakes and get back on track when errors occur, which limits their overall navigation performance. This deficiency raises an important question - \textbf{Can we teach robots to self-correct errors during navigation?}

For this problem, we analyze what kinds of errors to correct and how to teach the navigation model to correct them. As a Vision-Language-Action (VLA) task, VLN requires the model to dynamically perceive the environment and follow the given instruction to navigate. Errors often come from two sources:  misperception of landmarks and misunderstanding of instruction-specified actions. These errors propagate through the decision-making pipeline, adversely impacting movement prediction. Therefore, attention should be directed toward errors stemming from perception and actions. Besides, real-world applications impose time requirements on the model inference, necessitating that self-correction capabilities should be implicitly integrated into the model through training, rather than being achieved by increasing modules or the reasoning process. 

Consequently, we propose Self-correction Flywheel, a novel post-training paradigm for navigation. This approach stems from our observation that well-trained navigation models still produce error trajectories when evaluated on the training set. Rather than viewing these errors as mere shortcomings, we regard them as valuable opportunities to enhance the model further. Our Self-correction Flywheel proceeds through the following four steps: \textbf{(1)} Evaluating the trained model on its training set to collect error trajectories. \textbf{(2)} Then, we design an automatic approach capable of detecting the deviations and pinpointing their exact locations in error trajectories. \textbf{(3)} After identifying deviations, we create self-correction data from action and perception perspectives. For action correction, we gather trajectories that demonstrate effective recovery from deviations. For perception correction, we leverage large-scale multimodal models to analyze keyframes associated with navigation errors. \textbf{(4)} With these self-correction data, we drive the continued training of the navigation model to improve its performance. Completing the above four steps constitutes one round of the Self-correction Flywheel. When we continue to evaluate the model, which has undergone one round of self-correction training, on the training set, a remarkable thing happens. We can identify new error trajectories, thereby generating fresh self-correction data and further training the model. At this time, the Self-correction Flywheel is in motion, and the performance of the navigation model will continuously improve with multiple rounds of training iterations.

Furthermore, we design a suite of navigation fine-tuning strategies, including observation randomization, instruction generation, and general multimodal data recall. Through our proposed fine-tuning and post-training strategies, we develop a new monocular RGB-based VLA navigation model \method. On VLN-CE benchmarks R2R-CE and RxR-CE, \method achieves success rates of 65.1\% and 69.3\%, surpassing previous state-of-the-art models by 8.2\% and 16.4\%. The real robot tests conducted in diverse indoor and outdoor environments demonstrate that \method possesses strong capabilities of error correction, dynamic obstacle avoidance, and long instruction following, outperforming existing navigation models. 

\section{Related Work}
\subsection{Vision-and-Language Navigation}
Vision-and-Language Navigation (VLN) involves an embodied agent navigating to a target location following natural language instructions. Datasets like R2R~\cite{r2r} and RxR~\cite{rxr} provide navigation instructions and trajectories in the discretized MP3D~\cite{mp3d} environment, while VLN-CE~\cite{krantz2020beyond} adapts these to continuous settings. Current VLN-CE models can be categorized into two groups: topology graph-based approaches, such as BEVbert~\cite{an2023bevbert} and ETPnav~\cite{an2024etpnav}, which rely on multiple sensors to predict waypoints; and models built on pretrained vision-language models (VLMs), including NaVid~\cite{zhang2024navid}, Uni-NaVid~\cite{uninavid}, and NAVILA~\cite{cheng2024navila}, which infer actions end-to-end according to RGB observation. Existing methods commonly employ techniques such as auxiliary tasks~\cite{uninavid,zhang2024navid}, instruction augmentation~\cite{wei2025unseenseenrewritingobservationinstruction}, and dataset expansion~\cite{streamvln} to enhance performance, but devote less attention to error correction. For easier real robot application, we also construct our \method based on a pretrained VLM. However, we highlight the value of error correction, which helps us break through the performance bottleneck of current technologies. 

\subsection{Error Correction in Embodied Intelligence}
Errors are usually inevitable in embodied intelligence tasks. To enhance the robustness, the ability to correct errors is essential. Error correction methods have been explored in manipulation tasks~\cite{ha2023scalingdistillingdownlanguageguided,ma2023vipuniversalvisualreward,duan2024aha,liu2023reflect}. However, error correction in navigation tasks is less explored. SmartWay~\cite{SmartWay} uses closed-source large models to reflect on trajectories and decide whether to backtrack, while EnvolveNav~\cite{lin2025evolve} trains models to generate time-consuming chains of thought with limited improvement. These methods often require additional models or inference steps, reducing efficiency and hindering deployment in the real world. In contrast, our method implicitly teaches error correction through Self-correction Flywheel training, eliminating the need for additional modules or long thinking, thereby facilitating deployment on real robots.

\section{\method Model}
\subsection{Task Definition}
Given a language instruction $L_{nav}$, the vision-and-language navigation task requires the model to predict the next navigation action $a_{t+1} \in A$ at time step $t$ based on observation \(\{O_1, O_2, \ldots, O_t\}\). Recently, to overcome the reliance on multi-sensor, researchers~\citep{zhang2024navid} have simplified observation into a sequence of monocular RGB images $\{I_1, I_2 ... I_t\}$ captured during navigation. 

\begin{algorithm}
\caption{Self-correction Flywheel Post-training}
\label{alg:combined}
\KwIn{oracle trajectories $\{ T_g^{(i)}, L_{nav}^{(i)} \}$, dataset $D_{nav}$, model $\mathcal{M}$, number of flywheel iteration $N$, distance threshold $S$, trajectory planner $\Gamma$}
\KwOut{Model $\mathcal{M}$}

$\mathcal{M} \leftarrow \text{Train}(D_{nav},\mathcal{M})$\;

\For{$c \leftarrow 1$ \KwTo $N$}{
    $\{ T_m^{(i)} \} \leftarrow \mathcal{M} ( \{L_{nav}^{(i)}\} )$\;
    $D_{new} \leftarrow \emptyset$\;
    
    \For{each sample $i$ in the dataset}{
        $K^{(i)}, T_c^{(i)} \leftarrow \text{DeviDetect}(T_g^{(i)}, T_m^{(i)}, S, \Gamma)$\;
        $\text{Cap}^{(i)} \leftarrow \text{MLLM\_Description}( K^{(i)} )$\;
        $\text{Qa}^{(i)} \leftarrow \text{MLLM\_QA}( K^{(i)} )$\;
        Add $( T_c^{(i)}, \text{Cap}^{(i)}, \text{Qa}^{(i)} )$ to $D_{new}$\;
    }
    $D_{train} \leftarrow \text{Sample}( D_{nav} ) \cup \text{Sample}(D_{new})$\;
    $\mathcal{M} \leftarrow \text{Train}( D_{train},\mathcal{M})$\;
}
\end{algorithm}

\begin{figure*}[htp]
    \centering
    \includegraphics[width=0.94\linewidth]{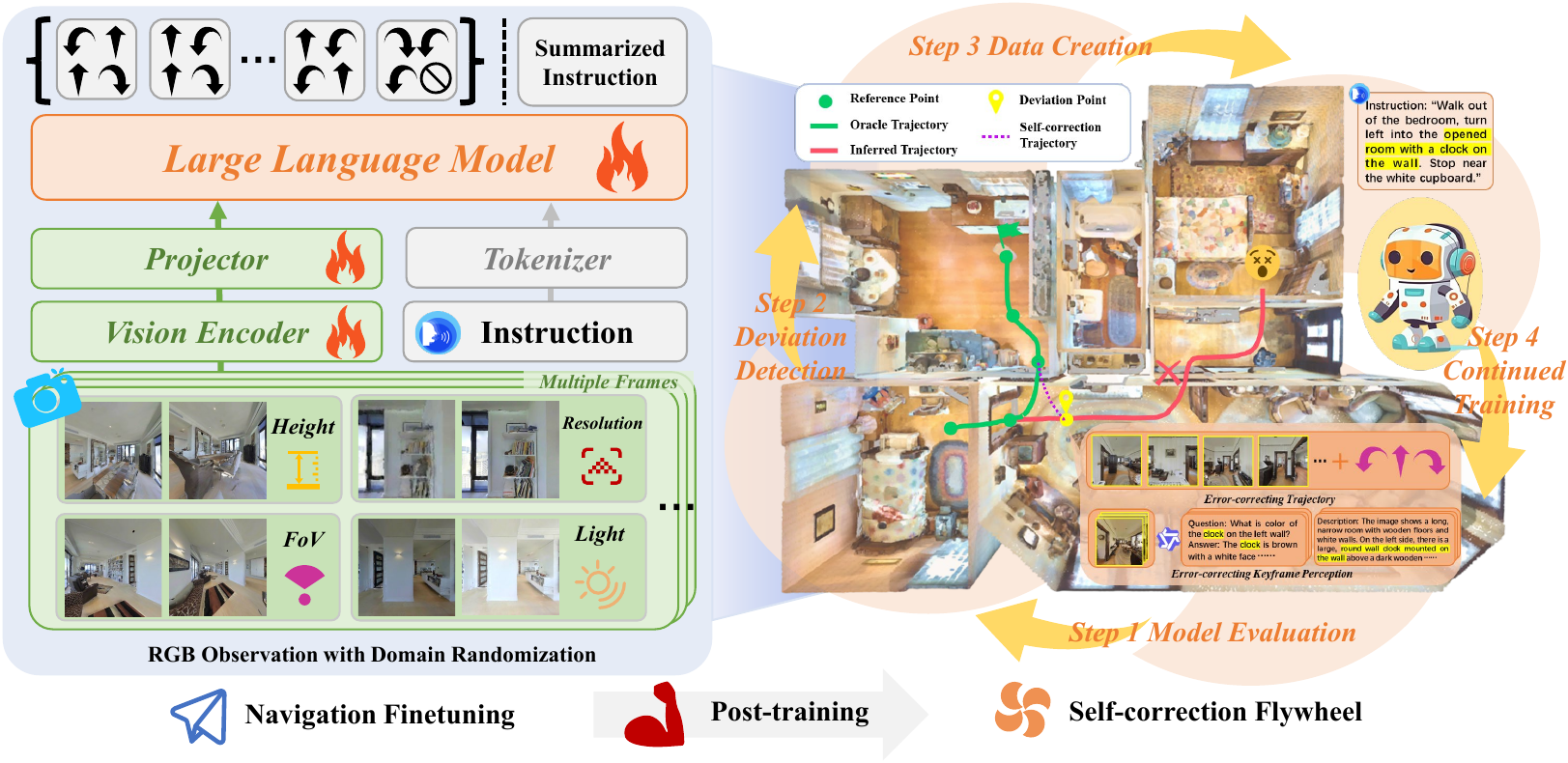}
    \vspace{-0.2cm}
    \caption{\textbf{The overview of \method training.} \method is first finetuned on the navigation tasks (Left), including action prediction and instruction generation. To enhance vision diversity, we implement a suite of domain randomization strategies. Subsequently, \method is post-trained with our proposed Self-correction Flywheel paradigm (Right). This paradigm operates in a continuous loop of \emph{model evaluation}, \emph{deviation detection}, \emph{data creation}, and \emph{continued training}. Specifically, the data creation part can automatically collect error-correcting trajectory and keyframe perception data. Through multiple training iterations, \method can learn how to recover from deviations.}
    \label{fig:correctnav}
    \vspace{-0.3cm}
\end{figure*}

\subsection{Model Structure}
Our \method~consists of three modules: the Vision Encoder \( v(\cdot) \), the Projector \( p(\cdot) \), and the Large Language Model (LLM) \( f(\cdot) \). Specifically, we employ SigLIP~\cite{zhai2023sigmoid}, a 2-layer MLP~\cite{liu2024improved}, and Qwen2~\cite{qwen2}. Given an RGB video, the Vision Encoder extracts visual features from the sampled frames, producing $Z_v = v(\{I_1, I_2 ... I_t\})$. The MLP Projector maps these visual features to the semantic space of the LLM, resulting in a sequence of visual tokens $H_v = p(Z_v)$. Using the visual tokens \( H_v \) together with textual tokens \( X \) encoded from the task instruction \( L \), the LLM \( f(\cdot) \) makes predictions in an auto-regressive manner. Before navigation finetuning, \method is initialized from LLaVA-Video 7B~\cite{178k}.

\subsection{Navigation Fine-tuning}
\subsubsection{Navigation Action Prediction}
We collect oracle navigation trajectories from VLN-CE R2R and RxR train splits in MP3D indoor scenes. Each oracle trajectory contains one navigation instruction and step-wise RGB observations and navigation actions \(\tau = (L_{nav}, \{(I_t, a_t)\}_{t=1}^T)\). To enhance visual diversity, we implement a set of domain randomization strategies. These strategies encompass randomizing camera height, adjusting the field of view, varying observation resolution, and altering illumination conditions, as shown in Figure~\ref{fig:correctnav}. With these strategies, we collected more than 2.1 million step-wise navigation action prediction data $D_{nav}$, including 527K samples from R2R and 1.58 million samples from RxR. In this task, we take navigation instruction $L_{nav}$ and step-wise RGB observations $\{I_1, I_2 ... I_t\}$ as \method's input and require the model to predict an action trunk $\{a_{t+1}, a_{t+2} ... a_{t+m}\}$ with m steps.

\begin{table*}[t]
    \small
    \centering    
    \setlength{\tabcolsep}{7.0pt}
    \scalebox{0.88}{
{\fontsize{8pt}{9pt}\selectfont
\begin{tabular}{lcccclcccclcccc}
\toprule
& \multicolumn{4}{c}{Observation} & & \multicolumn{4}{c}{R2R-CE Val-Unseen} & & \multicolumn{4}{c}{RxR-CE Val-Unseen} \\
\cmidrule(lr){2-5} \cmidrule(lr){7-10} \cmidrule(lr){12-15}
& S.RGB & Pano. & Depth & Odo. & & NE $\downarrow$ & OS $\uparrow$ & SR $\uparrow$ & SPL $\uparrow$ & & NE $\downarrow$ & SR $\uparrow$ & SPL $\uparrow$ & nDTW $\uparrow$ \\
\midrule
HPN+DN$^{*}$~\citep{krantz2021waypoint} &  & \checkmark & \checkmark & \checkmark & & 6.31 & 40.0 & 36.0 & 34.0 & & - & - & - & - \\
CMA$^{*}$~\citep{hong2022bridging} & & \checkmark & \checkmark & \checkmark & & 6.20 & 52.0 & 41.0 & 36.0 & & 8.76 & 26.5 & 22.1 & 47.0 \\
VLN$\circlearrowright$BERT$^{*}$~\citep{hong2022bridging} & & \checkmark & \checkmark & \checkmark & & 5.74 & 53.0 & 44.0 & 39.0 & & 8.98 & 27.0 & 22.6 & 46.7 \\
Sim2Sim$^{*}$~\citep{krantz2022sim} & & \checkmark  & \checkmark & \checkmark & & 6.07 & 52.0 & 43.0 & 36.0 & & - & - & - & - \\
GridMM$^{*}$~\citep{wang2023gridmm} & & \checkmark  & \checkmark & \checkmark & & 5.11 & 61.0 & 49.0 & 41.0 & & - & - & - & - \\
Ego$^{2}$-Map$^{*}$~\citep{hong2023learning} & & \checkmark  & \checkmark & \checkmark & & 5.54 & 56.0 & 47.0 & 41.0 & & - & - & - & - \\
DreamWalker$^{*}$~\citep{wang2023dreamwalker} & & \checkmark  & \checkmark & \checkmark & & 5.53 & 59.0 & 49.0 & 44.0 & & - & - & - & - \\
Reborn$^{*}$~\citep{an20221st} & & \checkmark  & \checkmark & \checkmark & & 5.40 & 57.0 & 50.0 & 46.0 & & 5.98 & 48.6 & 42.0 & 63.3 \\
ETPNav$^{*}$~\citep{an2024etpnav} & & \checkmark  & \checkmark & \checkmark & & 4.71 & 65.0 & 57.0 & 49.0 & & 5.64 & 54.7 & 44.8 & 61.9 \\
HNR$^{*}$~\citep{wang2024lookahead} & & \checkmark  & \checkmark & \checkmark & & 4.42 & 67.0 & 61.0 & 51.0 & & 5.50 & 56.3 & 46.7 & 63.5 \\
BEVBert$^{*}$~\citep{an2023bevbert} & & \checkmark  & \checkmark & \checkmark & & 4.57 & 67.0 & 59.0 & 50.0 & & - & - & - & - \\
HAMT+ScaleVLN$^{*}$~\citep{wang2023scaling} & & \checkmark  & \checkmark & \checkmark & & 4.80 & - & 55.0 & 51.0 & & - & - & - & - \\
\midrule
AG-CMTP~\citep{chen2021topological} & & \checkmark  & \checkmark & \checkmark & & 7.90 & 39.0 & 23.0 & 19.0 & & - & - & - & - \\
R2R-CMTP~\citep{chen2021topological} & & \checkmark  & \checkmark & \checkmark & & 7.90 & 38.0 & 26.0 & 22.0 & & - & - & - & - \\
InstructNav~\citep{long2024instructnavzeroshotgenericinstruction} & & \checkmark  & \checkmark & \checkmark & & 6.89 & - & 31.0 & 24.0 & & - & - & - & - \\
LAW~\citep{raychaudhuri2021language} & \checkmark & & \checkmark & \checkmark & & 6.83 & 44.0 & 35.0 & 31.0 & & 10.90 & 8.0 & 8.0 & 38.0 \\
CM2~\citep{georgakis2022cross} & \checkmark & & \checkmark & \checkmark & & 7.02 & 41.0 & 34.0 & 27.0 & & - & - & - & - \\
WS-MGMap~\citep{chen2022weakly} & \checkmark & & \checkmark & \checkmark & & 6.28 & 47.0 & 38.0 & 34.0 & & - & - & - & - \\
AO-Planner~\citep{chen2024affordances} & & \checkmark & \checkmark & & & 5.55 & 59.0 & 47.0 & 33.0 & & 7.06 & 43.3 & 30.5 & 50.1 \\
Seq2Seq~\citep{krantz2020beyond} & \checkmark & & \checkmark & & & 7.77 & 37.0 & 25.0 & 22.0 & & 12.10 & 13.9 & 11.9 & 30.8 \\
CMA~\citep{krantz2020beyond} & \checkmark & & \checkmark & & & 7.37 & 40.0 & 32.0 & 30.0 & & - & - & - & - \\
RGB-Seq2Seq~\citep{krantz2020beyond} & \checkmark & & & & & 10.10 & 8.0 & 0.0 & 0.0 & & - & - & - & - \\
RGB-CMA~\citep{krantz2020beyond} & \checkmark & & & & &  9.55 & 10.0 & 5.0 & 4.0 & & - & - & - & - \\
NaVid~\citep{zhang2024navid} & \checkmark & & & & &  5.47 & 49.0 & 37.0 & 35.0 & & - & - & - & - \\
Uni-NaVid~\citep{uninavid} & \checkmark  & & & & & 5.58 & 53.5 & 47.0 & 42.7 & & 6.24 & 48.7 & 40.9 & - \\
NaVILA~\citep{cheng2024navila} & \checkmark  & & & & & 5.22 & 62.5 & 54.0 & 49.0 & & 6.77 & 49.3 & 44.0 & 58.8 \\
StreamVLN~\citep{streamvln} & \checkmark  & & & & & 4.98 & 64.2 & 56.9 & 51.9 & & 6.22 & 52.9 & 46.0 & 61.9 \\
\rowcolor{gray!15}
\textbf{\method (Ours)} & \checkmark  & & & & & \bf4.24 & \bf67.5 & \bf65.1 & \bf62.3 & & \bf4.09 & \bf69.3 & \bf63.3 & \bf75.2 \\
\bottomrule
\end{tabular}}}
\vspace{-0.1cm}
\caption{\textbf{Comparison with state-of-the-art methods on the Val-Unseen split of R2R-CE~\citep{r2r} and RxR-CE~\citep{rxr}.} $^{*}$ indicates methods using the waypoint predictor from~\citet{hong2022bridging}. \method outperforms all methods that do not rely on simulator pre-trained waypoint predictors, even when those methods leverage additional inputs such as depth, panoramic views, and odometry.}
\label{tab:r2r_rxr}
\vspace{-0.3cm}
\end{table*}

\subsubsection{Trajectory-based Instruction Generation}
In this task, we collect complete oracle navigation trajectories from VLN-CE R2R and RxR datasets. Among these trajectories, 10K are from R2R and 20K are from RxR. \method needs to generate language-format navigation instructions based on the monocular RGB observation history. During the training, we input the RGB observations of the whole oracle trajectory $\{I_1, I_2 ... I_T\}$, and take the corresponding instruction $L_{nav}$ as the target.

\subsubsection{General Multimodal Data Recall}
The format of our downstream navigation task differs significantly from general multimodal training tasks. Only training on the navigation tasks results in general multimodal capability forgetting during the training. To address this, we include a subset of video data from the LLaVA-Video 178K dataset \cite{178k}. We focus on Activitynet-QA~\cite{yu2019activityqa} and NextQA~\cite{xiao2021next}, which emphasize temporal and spatial scene understanding, aligning with our goals. Therefore, we randomly sample 240K training instances from ActivityQA and NextQA to maintain the model’s general multimodal abilities.

\subsection{Self-correction Flywheel Post-training}
To teach the navigation model how to recover from deviations, we propose a new post-training paradigm, Self-correction Flywheel. One iteration of training includes model evaluation, deviation detection, self-correction data creation, and continued training. These four steps can form a closed loop at both ends to create a self-correcting flywheel. Through multiple training iterations, self-correction capabilities can be specifically improved. The overview is introduced in Algorithm~\ref{alg:combined}. Each step will be detailed below. 

\subsubsection{Step 1 - Model Evaluation on Train Split}
The training splits of R2R-CE and RxR-CE provide a large number of instructions and oracle trajectory pairs. In the dataset, each oracle trajectory is defined by a sequence of ordered reference points, denoted as $T_g =(G_1,\ldots, G_n)$. During the navigation fine-tuning, we have already used this data to provide step-by-step supervision signals for \method training. Although the model has been trained on these data, we found that it still makes errors when evaluated on the training set. We realize that this is an excellent source for collecting correction data. The training dataset not only contains abundant data but also includes ground truth reference points. Therefore, we collect error trajectories produced during model evaluation on the training set. These trajectories can be denoted by $T_m = (M_1, \ldots, M_m)$, where \( M_i \) represents the position of the robot at the \( i \) timestep.

\subsubsection{Step 2 - Trajectory Deviation Detection}
Since the collected error trajectories lack annotations indicating where deviations occur, we develop a method to detect such deviations. The key principle is to assess deviations by measuring the distance between the error trajectories and the oracle trajectories. To compute the distance from a robot position \( M_i \) to the oracle trajectory \( T_g \), we begin by uniformly interpolating between reference points, which forms an evenly spaced sequence $T_g'$. For each robot position \( M_i \in T_m \), we define the distance from \( M_i \) to the \( T_g \) as
\[
h_i = \min_{x \in T_g'} \| M_i - x \|_2
\]
We further define the orthogonal foot of \( M_i \) on $T_g$ as
\[
P_i = \operatorname*{arg\,min}_{P \in T_g'} \| M_i - P \|_2.
\]
Let \( S \) be a predefined threshold. If there exists a timestep \( t \) such that
\[h_t > S \quad \text{and} \quad h_i \leq S, \quad \forall i < t, i\in N^*\]
Then we claim that the model begins to deviate from the oracle trajectory at \( M_t \). The observations near timestep \( t \) can be marked as keyframes for error correction.

\subsubsection{Step 3 - Self-correction Data creation}
By analyzing deviations in error trajectories, we identify that navigation errors primarily originate from perception and action. Accordingly, we propose self-correction tasks and data creation methods addressing these two aspects. 

\noindent\textbf{Error-correcting Trajectory} To teach the model how to recover from deviations, we collect error-correcting trajectories based on the detected deviations. Given an oracle trajectory $T_g$ and model trajectory $T_m$ with deviations, we already detect the deviation point $M_t$ and the corresponding orthogonal foot \( P_t \) in Step 2. If $P_t$ lies on the segment \( \overline{G_k G_{k+1}}\) ($G_k, G_{k+1} \in T_g$), we can know the model has correctly passed through \( G_k \)  and all previous reference points but deviates slightly while moving towards \( G_{k+1} \). We then utilize a trajectory planner \( \Gamma \) to generate a new trajectory 
\[
T_e = (M_t,\, G_{k+1},\, \ldots,\, G_n)
\]
This trajectory begins at \( M_t \), passes through the subsequent reference points, and concludes at the destination \( G_n \). Thus, we get an error-correcting trajectory, which can serve as training data for action correction. The training is similar to navigation action prediction. To ensure the model focuses on learning correction behavior, action learning is only performed on the error-correcting trajectory $T_e$ while the trajectory before $M_t$ solely provides observation history.

\noindent\textbf{Keyframe Perception} To truly equip \method with error-correcting ability, we should not only teach it what to act but also why. Errors in the process of vision-language navigation often stem from multimodal perception errors made by the navigation model near the deviation position $M_t$. To enhance the multimodal perception capabilities of \method during correction training, we select the observation frame at $M_t$, as well as the frames before and after $M_t$, as correction keyframes $\{K_1, K_2, K_3\}$. We then leverage a multimodal LLM Qwen-VL-Plus to create vision analysis data based on these key correction frames as shown in the right part of Figure~\ref{fig:correctnav}. The first type of vision analysis is describing potential navigation landmarks, such as furniture, decorations, or architectural structures, that appear in the given frames. 
\[
C_i = \mathrm{MLLM}(K_i,\, L_{cap})
\]
The second type of vision analysis is generating visual question answering pairs about the frames. These questions concentrate on important visual elements in navigation, including object relative position, object color, and the current robot's orientation.
\[
\{(Q_j, A_j)\}_{j=1}^x = \mathrm{MLLM}(K_i,\, L_{qa})
\]
During the training, we input the observation video $\{I_1, I_2 ... K_i\}$ and use the caption $C$ as the target, train \method to comprehend the current observation; With the same video, and for any $(Q_i, A_i)$, we instruct \method to answer $Q_i$ based on current observation (\(K_i\)), activating \method to comprehend error correction behavior. 

\subsubsection{Step 4 - Model Continued Training}
With collected self-correction data, we continue the training of \method. To enhance efficiency, we randomly sample half of the error-correcting trajectories along with their corresponding keyframe perception data for training. Additionally, we incorporate oracle trajectories from the original training data to maintain training stability. The number of these oracle trajectories is set to be half the number of the sampled error-correcting trajectories. By leveraging these automatically generated data, we can further train \method to enhance its self-correction capabilities. At this time, we have completed one round of Self-correction Flywheel training. 

\subsubsection{Multi-Round Self-Correction Iteration}
When we test the self-corrected \method on the training set again, new error trajectories emerge. These new errors allow us to generate fresh correction task data for the continued training of \method. This starts the Self-correction Flywheel, enabling multiple rounds of self-correction training iterations. 

\subsection{Implementation Details}
\method is trained on a server with 8 NVIDIA A100 GPUs. The navigation finetuning requires 80 hours while one iteration of Self-correction Flywheel consumes 20 hours. At inference time, \method takes 16 sampled RGB frames as input and predicts an action chunk with 4 effective actions. 

\section{Experiments}
We conduct experiments to answer the questions:
(1) How does \method compare to state-of-the-art models on VLN-CE benchmarks? (2) What improvements does \method achieve through Self-correction Flywheel iterations? (3) What is the individual impact of each self-correction training technique on \method’s performance? (4) How effective is \method in the real world?

\begin{figure*}[htp]
    \centering
    \includegraphics[width=\linewidth]{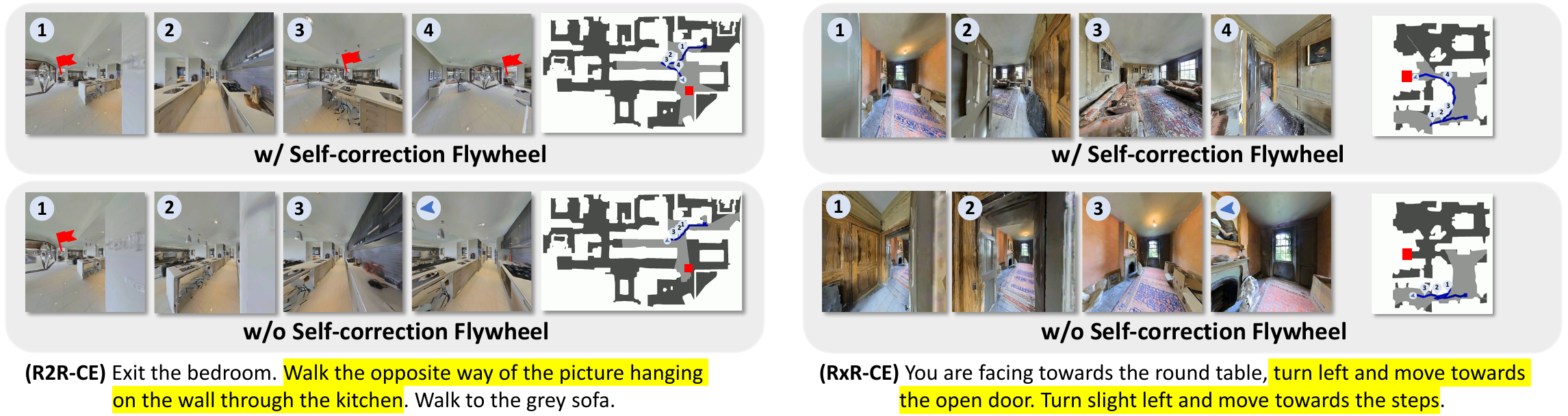}
    \vspace{-0.3cm}
    \caption{\textbf{Case study about \method with and without Self-correction Flywheel post-training.} Left Top: \method mistakenly enters the wrong path, loses the target, and then promptly turns back to return to the correct path. Right Top: \method first enters the front door, and after realizing there is no target (steps), it leaves and directly enters the correct side door. Vanilla \method fails in both cases.}
    \label{fig:comparison}
    \vspace{-0.3cm}
\end{figure*}

\subsection{Simulation Experiments}
\subsubsection{Environment and Metrics}
We evaluate our VLA on the VLN-CE benchmarks, which provide continuous environments for executing navigational actions in reconstructed photorealistic indoor scenes. We focus on the Val-Unseen split in both R2R (Room-to-Room) and RxR (Room-across-Room) datasets with VLN-CE, as these are the two most recognized benchmarks in VLN. Following the setting of VLN-CE~\cite{krantz2020beyond} benchmark, we take Habitat 3.0~\cite{puig2023habitat3} as the simulator to conduct the evaluation. Besides, we employ the following widely used evaluation metrics: Navigation Error (NE), Oracle Success Rate (OS), Success Rate (SR), Success-weighted Path Length (SPL), and normalized dynamic time wrapping (nDTW). Navigation Error, representing the average distance in meters between the agent’s final location and the target; Success Rate, indicating the proportion of paths with NE less than 3 meters; Oracle Success Rate, representing the SR given oracle stop policy. nDTW~\cite{ndtw} involves time warping to measure the distance between the model trajectory and ground truth.

\subsubsection{Comparison with other VLN-CE Models}
We compare our VLA with existing VLN-CE models on R2R-CE and RxR-CE benchmarks. These baselines include waypoint predictor-based models and navigation large models. From Table~\ref{tab:r2r_rxr}, although our \method only takes monocular RGB observation as input, it outperforms all existing models on R2R-CE and RxR-CE benchmarks. Compared to the state-of-the-art navigation large model StreamVLN, \method achieves an improvement of 8.2\% and 16.4\% in success rates on R2R-CE and RxR-CE, respectively. Additionally, \method outperforms the top waypoint predictor-based models, surpassing HNR by 4.1\% on R2R-CE and by 13.0\% on RxR-CE.

\subsubsection{The Effect of Self-correction Training Technologies}
To study the contribution of different self-correction training technologies to the improvement of model performance, we conducted ablation studies by removing each technology individually during the first iteration of the Self-correction Flywheel. As shown in Table~\ref{tab:ablation}, eliminating any of these technologies results in reduced performance of \method on both the R2R-CE and RxR-CE Val-Unseen splits. Notably, removing the Navigation Trajectory Correction strategy causes the most substantial performance drop. These ablation results confirm that each self-correction training technology we proposed contributes positively to enhancing the model’s overall effectiveness.

\subsubsection{The Power of Self-correction Flywheel Iteration}
\begin{table}[t]
    \small
    \centering    
    \setlength{\tabcolsep}{3pt}
    \scalebox{0.78}{
    {\fontsize{8pt}{9pt}\selectfont
    \begin{tabular}{lcccccccc}  
    \toprule
    & \multicolumn{3}{c}{R2R-CE Val-Unseen} & & \multicolumn{3}{c}{RxR-CE Val-Unseen} \\
    \cmidrule(lr){2-4} \cmidrule(lr){6-8}  
    & NE $\downarrow$ & SR $\uparrow$ & SPL $\uparrow$ & & NE $\downarrow$ & SR $\uparrow$ & SPL $\uparrow$ \\
    \midrule
    \rowcolor{gray!15}
    \textbf{\method}  & \textbf{4.50} & \textbf{63.0} & \textbf{59.0} & & \textbf{4.40} & \textbf{63.1} & \textbf{57.0} \\
    \quad w/o Navigation Trajectory Correction   & 4.92 & 59.2 & 57.2 & & 4.55 & 60.7 & 55.1 \\
    \quad w/o Error-correcting Keyframe Perception  & 4.70 & 60.1 & 56.5 & & 4.47 & 61.0 & 56.3 \\
    \quad w/o Data Sampling Strategy  & 4.71 & 60.0 & 57.5 & & 4.47 & 62.2 & 56.2 \\
    \bottomrule
    \end{tabular}
    }}
\caption{\textbf{Ablation study of self-correction flywheel post-training technologies on the Val-Unseen splits of R2R-CE~\citep{r2r} and RxR-CE~\citep{rxr}.} The experiments are conducted based on the \emph{$1_{st}$} self-correction flywheel post-training.}
\label{tab:ablation}
\end{table}

\begin{figure}[htp]
    \centering
    \includegraphics[width=\linewidth]{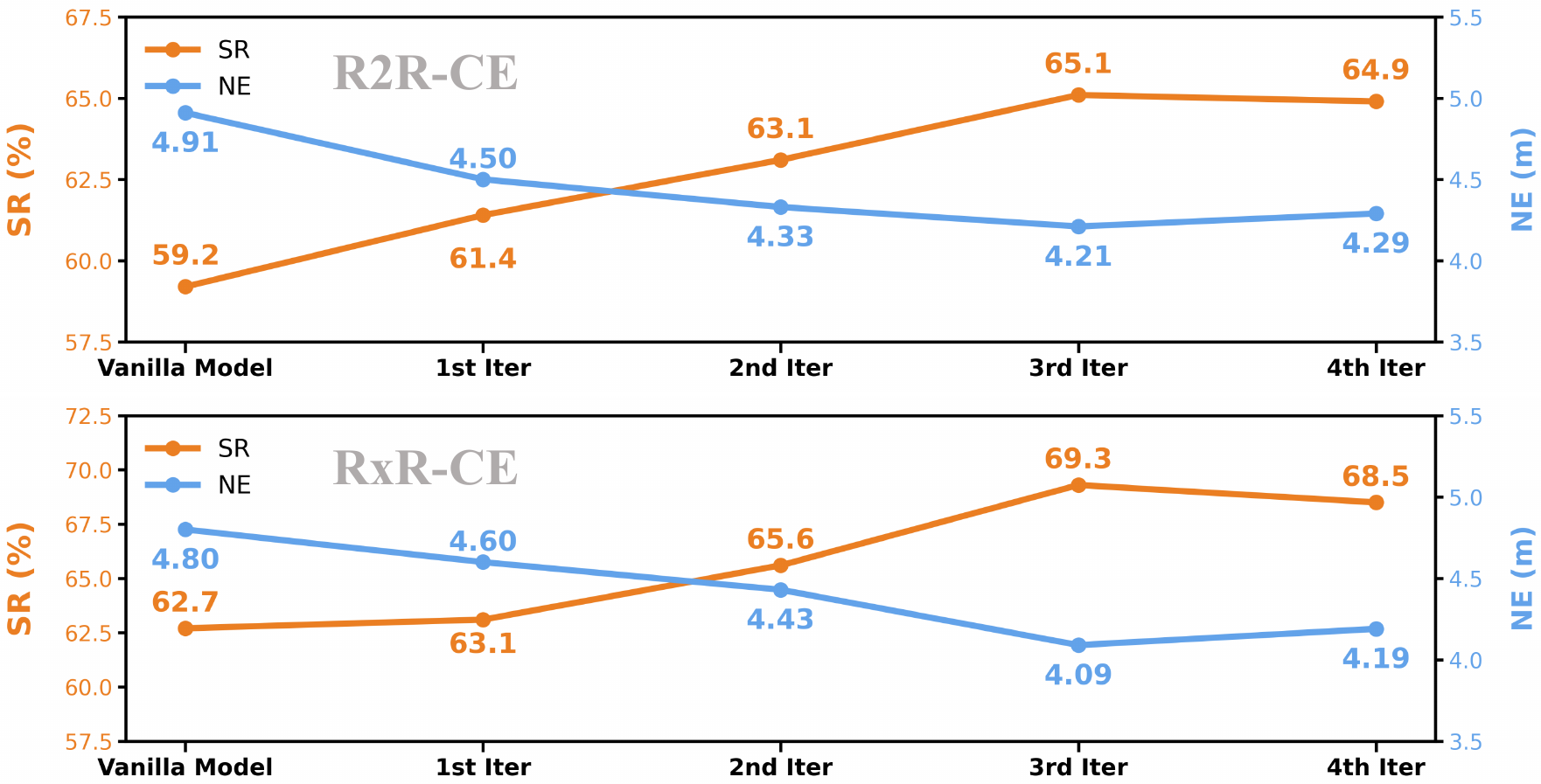}
    \caption{\textbf{\method's performance on R2R-CE and RxR-CE Val-Unseen splits over Self-correction Flywheel iterations.}}
    \label{fig:flywheel}
\end{figure}

To investigate the power of the Self-correction Flywheel on improving the navigation performance, we evaluate \method’s success rate and navigation errors on the R2R-CE and RxR-CE Val-Unseen benchmarks after each Self-correction Flywheel training iteration. The experimental results are plotted as the line graphs shown in Figure~\ref{fig:flywheel}. From the figure, we can observe that as the Self-correction Flywheel training iterations progress, \method demonstrates a continuous performance improvement on both benchmarks in the first three iterations. This quantitatively demonstrates that multiple iterations of the Self-correction Flywheel can effectively enhance the capabilities of the navigation VLA model. When \method's performance drops in the fourth iteration, we stop the training. Additionally, we conduct a qualitative analysis of the Self-correction Flywheel’s effects, as illustrated in Figure~\ref{fig:comparison}. From the figure, \method post-trained with Self-correction Flywheel achieves error correction capability compared with the vanilla \method. 

\begin{figure*}[htp]
    \centering 
    \includegraphics[width=\linewidth]{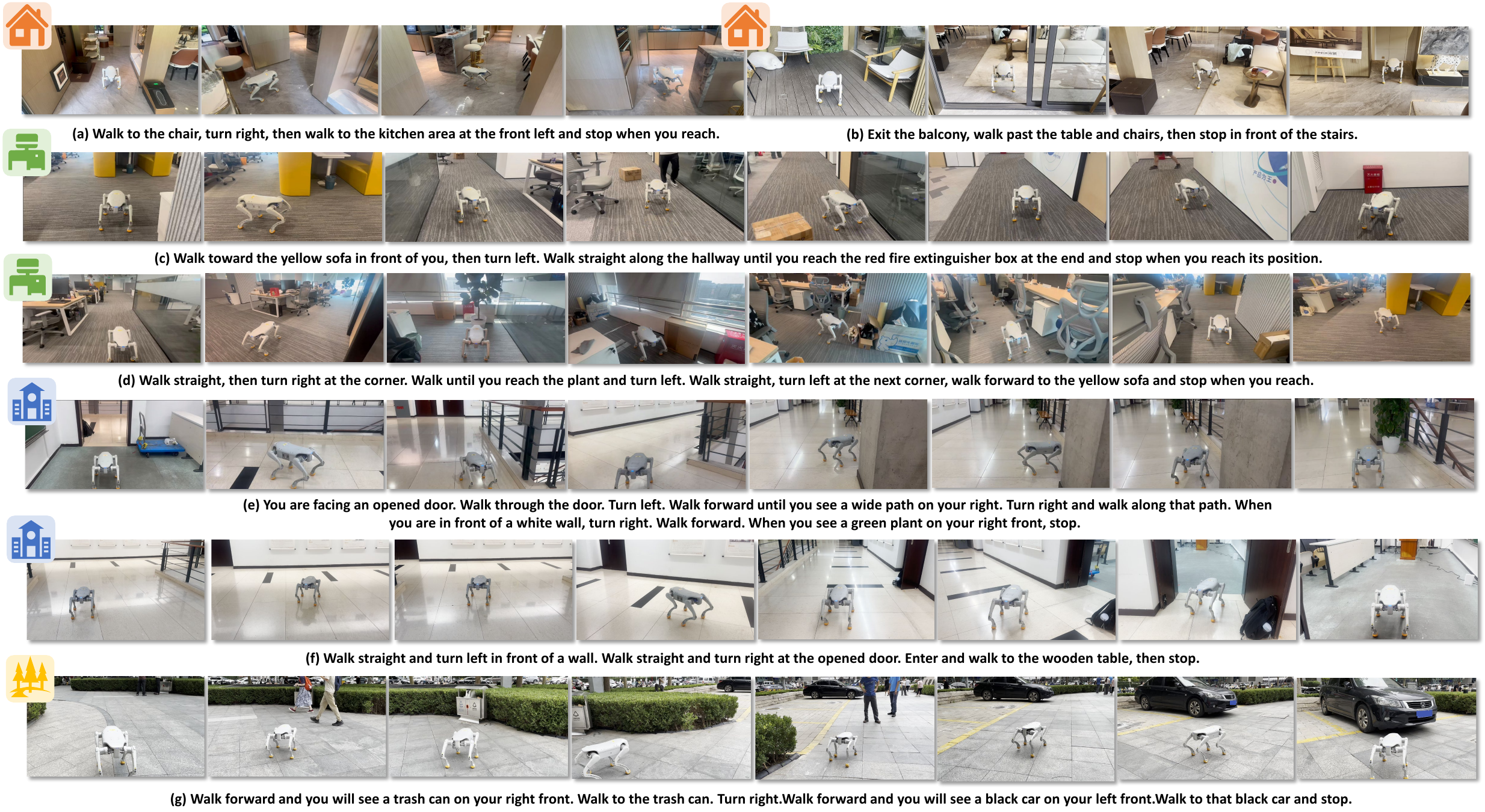}
    \captionof{figure}{\textbf{Qualitative results from the real-world deployment of \method.} (c)(d) The robot dynamically avoids pedestrians and obstacles, correctly passing through cluttered environments to reach the destination. (e)(f) The robot successfully recovers from a navigation error to complete a long-horizon instruction. (g) The robot completes outdoor long-distance navigation. Videos are shown on our project website.}
    \label{fig:real_robot}
\end{figure*}

\subsection{Real Robot Experiments}
\begin{table}[!t]
\small
\centering
\setlength{\tabcolsep}{1.5pt}
\scalebox{0.77}{
{\fontsize{8pt}{9pt}\selectfont
\begin{tabular}{l l cccc cccc cccc}
\toprule
& & \multicolumn{4}{c}{\texttt{Office}} & \multicolumn{4}{c}{\texttt{Home}} & \multicolumn{4}{c}{\texttt{Campus}} \\
\cmidrule(lr){3-6} \cmidrule(lr){7-10} \cmidrule(lr){11-14}
& & \multicolumn{2}{c}{Simple} & \multicolumn{2}{c}{Complex} & \multicolumn{2}{c}{Simple} & \multicolumn{2}{c}{Complex} & \multicolumn{2}{c}{Simple} & \multicolumn{2}{c}{Complex} \\
\cmidrule(lr){3-4} \cmidrule(lr){5-6} \cmidrule(lr){7-8} \cmidrule(lr){9-10} \cmidrule(lr){11-12} \cmidrule(lr){13-14}
& & NE$\downarrow$ & SR$\uparrow$ & NE$\downarrow$ & SR$\uparrow$ & NE$\downarrow$ & SR$\uparrow$ & NE$\downarrow$ & SR$\uparrow$ & NE$\downarrow$ & SR$\uparrow$ & NE$\downarrow$ & SR$\uparrow$ \\ 
\midrule
NaVid~\citep{zhang2024navid} & & 1.88 & 0.55  & 4.89 & 0.30 & 2.22 & 0.50 & 5.27 & 0.15 & 1.94 & 0.55 & 5.02 & 0.25 \\
NaVILA~\citep{cheng2024navila} & & 2.06 & 0.45 & 5.25 & 0.20 & 2.21 & 0.45 & 5.49 & 0.10 & 1.97 & 0.50 & 5.18 &  0.20 \\
\rowcolor{gray!15}
\textbf{\method (Ours)} & & \textbf{1.52} & \textbf{0.80} & \textbf{1.81} & \textbf{0.75} & \textbf{1.33} & \textbf{0.95} & \textbf{1.54} & \textbf{0.80} & \textbf{1.47} & \textbf{0.85} & \textbf{1.86} & \textbf{0.75} \\
\bottomrule
\end{tabular}
}
}
\caption{\label{tab:real_eval} \textbf{Real-world experiments in different environments.} Simple and Complex refer to simple and complex instruction-following tasks, respectively. }
\end{table}

For real-world experiments, we use the AgiBot Lingxi D1 quadruped robot as our platform. Each Lingxi D1 robot is equipped with a monocular RGB camera and robust motion APIs. After the robot receives a navigation instruction, it will upload the RGB observation image to the \method deployed on the remote server with an NVIDIA A100 GPU. \method will predict the navigation action chunk with four actions and call the D1 motion API to execute.

To comprehensively evaluate the effectiveness of our approach, we conduct comparisons with two state-of-the-art navigation large models, NaVID and NaVILA in the office, home, and campus. In each scenario, we test every model on 20 simple instructions and 20 complex instructions, respectively. Complex instructions involve long trajectories, complex architectural structures, crowded obstacles, and dynamic scene changes. The real-world quantitative performances in terms of Success Rate (SR) and Navigation Error (NE) metrics are reported in Table~\ref{tab:real_eval}. Figure~\ref{fig:real_robot} demonstrates the real-world qualitative performances of \method in indoor and outdoor environments.

From Table~\ref{tab:real_eval}, we can observe that compared to existing navigation large models, \method demonstrates a stronger ability to execute navigation instructions in the real world. As shown in Figure~\ref{fig:real_robot}, such improvement mainly stems from the deviation correction capabilities that \method acquires through Self-correction Flywheel post-training. These capabilities enhance the robustness of \method on complex instructions, enabling it to quickly correct its own errors or adapt to changes in the environment.

\section{Limitation and Future Work}
Although monocular RGB observation-based VLA navigation models, including our \method, save on the cost of additional sensors, they face a shared limitation: inadequate precision in perceiving the relative positional relationship between the robot’s body and its surroundings. The potential risk arising from this deficiency is that when the robot, such as a quadrupedal robot, passes close to obstacles, its hind legs may scrape against them. A promising direction for future research is to explore how to incorporate the robot’s body dimensions and state information as prior knowledge for navigation model inference, thereby further improving monocular RGB-based VLA navigation models.

\bibliography{aaai2026}

\end{document}